\documentclass[lettersize,journal]{IEEEtran}
\usepackage{amsmath,amsfonts}
\usepackage{algorithmic}
\usepackage{algorithm}
\usepackage{array}
\usepackage[caption=false,font=normalsize,labelfont=sf,textfont=sf]{subfig}
\usepackage{textcomp}
\usepackage{stfloats}
\usepackage{url}
\usepackage{verbatim}
\usepackage{graphicx}
\usepackage{cite}
\usepackage{enumitem}
\usepackage{multirow}
\usepackage{color}
\usepackage{graphicx}
\usepackage{booktabs}
\usepackage[table]{xcolor}
\begin{document}

\title{Reasoning in Action: MCTS-Driven Knowledge Retrieval for Large Language Models}

\author{Shuqi Liu\textsuperscript{1}, Bowei He\textsuperscript{1}, Chen Ma\textsuperscript{1}, Linqi Song\textsuperscript{1, 2, *\thanks{*Corresponding Author. This work was supported in part by the Research Grants Council of the Hong Kong SAR under Grant GRF 11217823 and Collaborative Research Fund C1042-23GF, the National Natural Science Foundation of China under Grant 62371411, InnoHK initiative, the Government of the HKSAR, Laboratory for AI-Powered Financial Technologies.}}\\
\textsuperscript{1}City University of Hong Kong\\
\textsuperscript{2}City University of Hong Kong Shenzhen Research Institute\\}


\markboth{Journal of \LaTeX\ Class Files,~Vol.~14, No.~8, August~2021}%
{Shell \MakeLowercase{\textit{et al.}}: A Sample Article Using IEEEtran.cls for IEEE Journals}


\maketitle

\begin{abstract}
Large language models (LLMs) typically enhance their performance through either the retrieval of semantically similar information or the improvement of their reasoning capabilities. However, a significant challenge remains in effectively integrating both retrieval and reasoning strategies to optimize LLM performance. In this paper, we introduce a reasoning-aware knowledge retrieval method that enriches LLMs with information aligned to the logical structure of conversations, moving beyond surface-level semantic similarity. We follow a coarse-to-fine approach for knowledge retrieval. First, we identify a contextually relevant sub-region of the knowledge base, ensuring that all sentences within it are relevant to the context topic. Next, we refine our search within this sub-region to extract knowledge that is specifically relevant to the reasoning process. Throughout both phases, we employ the Monte Carlo Tree Search-inspired search method to effectively navigate through knowledge sentences using common keywords. Experiments on two multi-turn dialogue datasets demonstrate that our knowledge retrieval approach not only aligns more closely with the underlying reasoning in human conversations but also significantly enhances the diversity of the retrieved knowledge, resulting in more informative and creative responses.
\end{abstract}


\section{Introduction}
\IEEEPARstart{L}{arge} language models (LLMs) have made significant strides in natural language processing, demonstrating impressive capabilities in generating coherent and contextually relevant responses \cite{touvron2023llama, ouyang2022training}. Their performance is often enhanced through two primary strategies: providing auxiliary information via retrieval \cite{pan2022knowledge, gopalakrishnan2023topical} and enhancing reasoning abilities within prompts \cite{wei2022chain}. 
Retrieval Augmented Generation (RAG) techniques focus on leverages embedding similarity searches to identify and incorporate information that is contextually relevant to a given query \cite{fan2024survey}. This integration not only improves the accuracy of generated outputs but also ensures that they are relevant and up-to-date.
However, these methods often focus primarily on retrieving directly relevant data, which may not fully capture the complexities of conversational logic and reasoning. 

Another way to enhance the responses of LLMs is by improving their reasoning abilities \cite{wei2022chain, havrillaglore}. Strong reasoning skills enable LLMs to achieve a deeper understanding of context and generate responses that logically connect to that context. While prompts can encourage LLMs to produce reasoning outcomes, structured reasoning resources, such as knowledge graphs, offer more accurate and controllable logical inferences for causal relationships \cite{kim2024causal}.
However, integrating retrieval and reasoning presents challenges; simply linking context to reasoning outcomes using current embedding similarity search methods is often ineffective.

To address this challenge, we present a reasoning-aware knowledge retriever that enhances LLMs with retrieved information aligned to the logical structure of conversations. 
We follow a coarse-to-fine approach for knowledge retrieval. First, we identify a contextually relevant sub-region of the knowledge base, ensuring that all sentences within it are relevant to the context topic. Next, we refine our search within this sub-region to extract knowledge that is specifically relevant to the reasoning process. 
In both phases, we utilize a Monte Carlo Tree Search (MCTS)-inspired method to effectively navigate the external knowledge base. We treat the sentences in the knowledge base as interconnected pieces of information, allowing us to traverse from one sentence to another through their common keywords.

\begin{figure*}[htbp]
\centering
\includegraphics[width=1.0\textwidth]{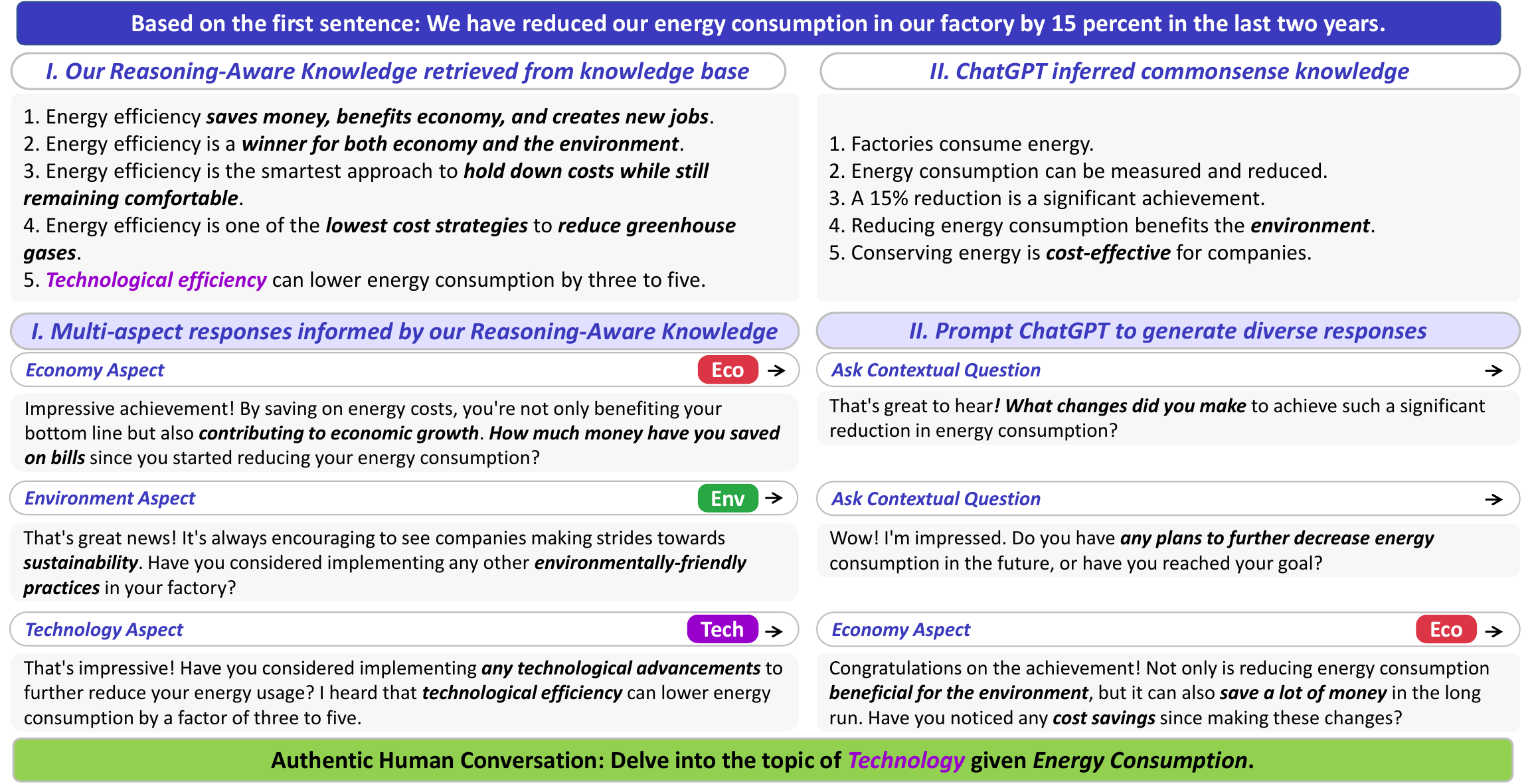}
\caption{
Illustration of the improved response quality achieved by integrating reasoning-aware knowledge into LLM. The retrieved knowledge exhibits higher pairwise diversity and aligns more closely with the logical structure of human conversations, resulting in more informative and multifaceted responses.
}
\label{figure:intro}
\end{figure*}

We use two representative multi-turn dialogue datasets as our evaluation benchmark. These datasets are suitable for our study as they require both relevant knowledge and reasoning to generate human-like responses. 
We systematically evaluate the impact of providing either semantically relevant knowledge or reasoning-aware knowledge to the LLM. Our findings confirm that the retrieved reasoning-aware knowledge aligns more closely with the logical structure of human conversations and exhibits higher pairwise diversity. As shown in Figure \ref{figure:intro}, by integrating reasoning-aware knowledge, LLMs can generate more informative and multifaceted responses.

\maketitle

\section{Related Works}
\noindent{\textbf{Knowledge Retrieval for Conversational Agents.}}
Recently, researchers have been designing methods to use Knowledge Base (KB) or Knowledge Graph (KG) as external knowledge to augment the response of conversational agents.
One line of works addressed the issue of selecting relevant facts from KBs or documents based on semantic similarity \cite{lewis2020retrieval, izacard2022few}. However, these approaches primarily focus on one-hop semantic searching and might not achieve appropriate knowledge selection. 
Others focus on incorporating the KG embeddings to improve text generation \cite{zhou2018commonsense, zhang2020grounded}. 
Beyond a simple concatenation, other works have explored fine-tuning the language models directly on knowledge graph triplets by transforming the KG triplets into readable sentences \cite{guan2020knowledge, moiseev-etal-2022-skill}. 
However, the binary relations in triple-based KGs limit the scope of encodable knowledge.
Instead, we leverage GenericsKB \cite{bhakthavatsalam2020genericskb}, which contains a wealth of commonsense sentences that capture the intricate relationships spanning across the sentences. 
Unlike previous document retrieval methods that select a single knowledge sentence, we employ an iterative search method to navigate through multiple sentences. 

\noindent{\textbf{Commonsense Reasoning for Conversational Agents.} }
Additionally, efforts have been made to leverage commonsense reasoning in conversational agents.
One line of the method is to create a custom dataset with annotations designed for learning commonsense \cite{ghosal2022cicero}.
The other line utilizes the prominent neural commonsense model COMET \cite{bosselut2019comet, hwang2021comet} to enhance conversational agents. 
This advancement has enabled researchers to generate novel commonsense explanations in the form of natural language, which are more flexible and extensible than knowledge graph-based methods.
These works either utilize COMET knowledge for emotion detection to generate more empathetic responses \cite{zhu2021topic, sabour2022cem, zhao2023knowledge} or enrich the understanding of the conversation by adding commonsense explanations obtained from COMET \cite{majumder-etal-2020-like, lin2022makes}.  
However, previous work ignores the fact that conversational agents require both reasoning ability and complex commonsense knowledge in the real world to surpass surface-level comprehension. Instead, we model our reasoning-aware commonsense retrieval method as a multi-objective optimization problem to consider reasoning and commonsense knowledge simultaneously. 

\noindent{\textbf{Augmented Language Models.}} 
Although LLMs have fueled dramatic progress in complex language benchmarks, they still face challenges in generating nonfactual but plausible hallucinations.
A growing research trend aims to solve the issues by augmenting LLMs with external tools. Augmented Language Models (ALMs) enhance Large Language Models (LLMs) by integrating external tools such as retrievers, search engines, calculators, and code interpreters \cite{mialon2023augmented}. 
Notable examples include LaMDA \cite{thoppilan2022lamda} and BlenderBot \cite{shuster2022blenderbot}, which incorporate search engines to facilitate richer, open-domain conversations. Similarly, WebGPT \cite{nakano2021webgpt} is designed to interact seamlessly with a web browser, enabling it to access and utilize up-to-date web information. 
Researchers also demonstrate that augmenting LLMs with reasoning and tools leads to better performance. The combination of reasoning and tools within LLMs should allow them to solve a broad range of complex tasks without heuristics, hence with better generalization capabilities.
In this paper, we take a step further by combining reasoning with a knowledge retriever to enable more informed and coherent responses in conversational agents.

\section{Methodology}

\begin{figure*}[htbp]
\centering
\includegraphics[width=1.0\textwidth]{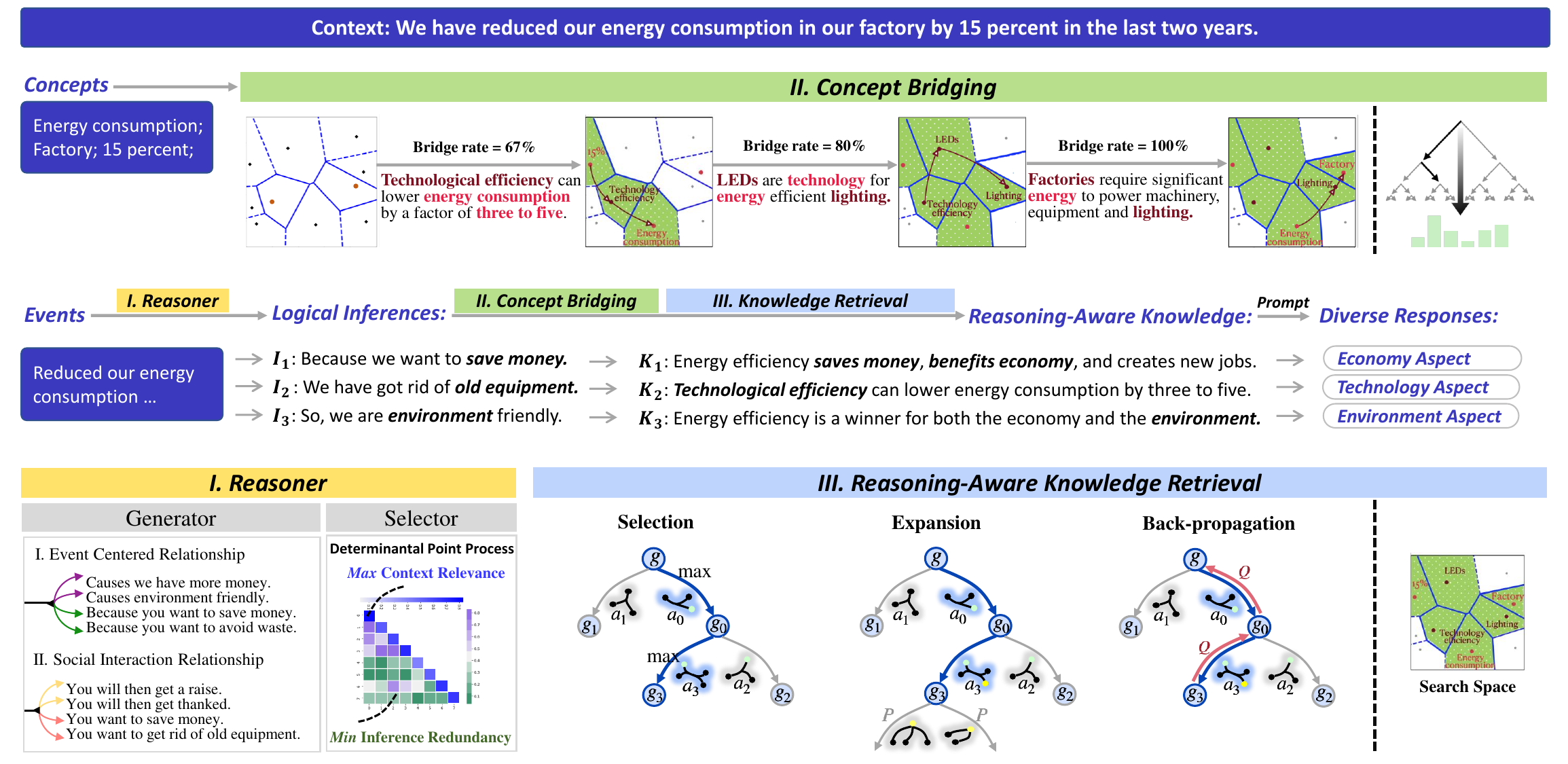}
\caption{Illustration of our Monte Carlo Tree Search (MCTS)-inspired reasoning-aware knowledge retriever, comprising three interconnected modules: (1) Reasoner, (2) Concept Bridging, and (3) Reasoning-Aware Knowledge Retrieval. The Concept Bridging and Reasoning-Aware Knowledge Retrieval modules utilize the MCTS-inspired knowledge retrieval method, each with its own task-specific policy and critic models.
}
\label{figure:method}
\end{figure*}

Our Monte Carlo Tree Search (MCTS)-inspired reasoning-aware knowledge retriever, illustrated in Figure \ref{figure:method}, consists of three interconnected modules: a Reasoner, a Concept Bridging module, and a Reasoning-Aware Knowledge Retrieval module. We first outline the overall retrieval process and then delve into the details of each module. Notably, both the Concept Bridging and Reasoning-Aware Knowledge Retrieval modules leverage the MCTS-inspired knowledge retrieval method, each with its own task-specific policy and critic models.

\subsection{Overview}

Our reasoning-aware knowledge retriever aims to traverse a commonsense knowledge base (CSKB) $K$ and retrieve a diverse set of knowledge sentences $k$ that reflect the underlying logical structure in a given conversation context $C$. To achieve this, we employ a two-step approach. First, the Diversity-Preserving Reasoner, denoted as $P_{r}$, generates a diverse set of logical inferences as reasoning outcomes $R$. Then, we formulate the problem of retrieving knowledge that aligns closely with both the conversation context and a specific logical inference as a multi-objective optimization (MOO) problem, denoted as $P_{o}$. The knowledge retrieved from the CSKB, denoted as $k$, serves as the decision variable.

To solve the MOO problem, we utilize the Epsilon-Constraint method \cite{mavrotas2009effective}, which converts one objective into a constraint with a predefined bound $\epsilon$, allowing us to focus on optimizing the remaining objective. In our case, we convert the context-coherence objective into the constraint by narrowing down the search space to a context-relevant sub-region ${K}_{c}$ using the Concept Bridging module $P_{b}$. This sub-region includes both explicitly mentioned concepts within the context and implicit concepts that serve as connectors. These concept connectors enrich the context-relevant sub-region and provide essential underlying information in the context.

The Reasoning-Aware Knowledge Retrieval module $P_{k}$ then focuses on extracting knowledge that directly supports the reasoning outcomes within this narrowed-down sub-region. A probabilistic view of the retrieval process is:

\begin{equation}
\begin{aligned}
    \max_{k\in {K}} P(k|C) &= P_{r}(R|C)\max_{k\in {K}}P_{o}(k|C, R) \\
    &= P_{r}(R|C) \max_{k\in {K}_{c}} P_{k}(k|R) \\  \qquad &  \textrm{s.t.}  \quad P_{b}({K}_{c}|C) \leq \epsilon \
\end{aligned}
\end{equation}

\subsection{Reasoner}

Our approach to generating a diverse set of logical inferences involves a two-step process. First, we employ COMET \cite{hwang2021comet}, denoted as $P_{r}$, to produce logical inference candidates ${R}{c}$, with a specific focus on event-centered and social-interaction relationships, as illustrated in Table \ref{tab:relation}. Each candidate inference $r \in {R}{c}$ is assigned a score $\tau_{r}$, which reflects the model's confidence in the inference $y_t$ with length $|d|$. This score is computed by concatenating the context $C$ and the commonsense relationship $rel$.

\begin{equation}
    \tau_{r} = \frac{1}{|d|} \sum_{t=1}^{|d|} P_{r}(y_{t}|y_{<t}, rel, C)
\end{equation}

To select a diverse subset of logical inferences ${R} \subset {R}_{c}$, we utilize the Determinantal Point Process algorithm \cite{kulesza2011k}. This selection process is based on a kernel matrix, where the diagonal entries represent the relevance scores of each inference to the dialogue context, while the off-diagonal entries quantify the redundancy between pairs of inferences \cite{song2020mpnet}. 

\begin{table}
\caption{Natural language statement for each relation used for commonsense reasoning model.}
    \centering
    \resizebox{0.47\textwidth}{!}{%
    \begin{tabular}{l>{\centering\arraybackslash}p{4.5cm}}
         \toprule
         \cellcolor{gray!20}Event Centered & \cellcolor{gray!20}Statement \\
         Causes & causes \\ 
         xReason & because  \\
         HinderedBy &  can be hindered by \\
         IsBefore &  happens before \\
         IsAfter & happens after \\ \hline
         \cellcolor{gray!20}Social Interaction & \cellcolor{gray!20}Statement \\
         xNeed & but before, x needed \\ 
         xAttr &  X is seen as \\
         xEffect &   as a result, x will\\
         xReact &  as a result, x feels  \\
         xWant &  as a result, x wants \\
         xIntent & because x wanted  \\ 
         \bottomrule
    \end{tabular}
    }
    \label{tab:relation}
\end{table}

\subsection{MCTS-inspired Knowledge Retrieval}

To effectively traverse the knowledge base and leverage complex connections between sentences, we formulate the knowledge retrieval process as a knowledge base walking process. This process moves from one knowledge sentence to another through a common concept node.

We denote a corpus of knowledge sentences as $K$, and use $N$ to represent the set of concept nodes. A knowledge sentence $k_{i} \in K$ describes generic world knowledge, and a concept $n_{j} \in N$ is a noun or noun phrase mentioned in these facts.

To facilitate the knowledge retrieval process, we expand the candidate knowledge sentences for each concept node $n_i$ by introducing a node group ${G}({n_{i})}\subset{N}$. This node group consists of other semantically relevant nodes around $n{i}$, and includes all knowledge sentences connected to the nodes in the group, denoted as $K_{{G}({n_{i})}} \subset {K}$. We construct these node groups by segmenting the entire set of concepts in the knowledge base based on semantic similarity, where each node group corresponds to a cluster that includes the node.


In our knowledge retrieval process, we formulate the problem as a Markov Decision Process (MDP) with states, actions, and transition probabilities. At each time step $t$, the state $s_t$ consists of three components: the communication context $C$, a knowledge sentence, and a marked concept node within the sentence that is most semantically similar to the context.
The action taken at each state $s_t$ involves selecting a knowledge sentence from the associated sentences of the current visiting concept node $K_{{G}({n_{t})}}$. The state then transits to the selected knowledge sentence, and one of the concepts is marked for the next iteration of expansion and selection. 

Since the transition probabilities are deterministic due to the static nature of the knowledge base, we can exploit this model-based feature and employ Monte Carlo Tree Search (MCTS) in the knowledge retrieval process. To optimize the speed of retrieval, we modify the traditional MCTS approach \cite{james2017analysis} by eliminating the rollout phase. Instead, the search process involves iterative steps of selection, expansion, evaluation, and back-propagation, gradually constructing a chain of knowledge sentences connected through common concepts. To guide our knowledge base traversal, we learn a policy model $\pi_{\theta }(a_{t}|s_{t})$ and a critic model $c_{\theta }(s_{t})$. The policy model allows us to sample knowledge sentences conditioned on a task-specific goal, while the critic model estimates our ability to find the knowledge sentences that satisfy the goal.

Our Concept Bridging and Reasoning-Aware Knowledge Retrieval modules both utilize the MCTS-inspired knowledge retrieval method. However, each module has its own task-specific policy and critic models, which we will describe in the following sections.

\subsection{Concept Bridging Module}

The Concept Bridging module aims to identify a subset of knowledge sentences that closely relate to the conversation context. This subset should cover both explicit concepts mentioned in the context and the implicit concepts necessary to connect them. Implicit concepts are crucial for filling information gaps and revealing a deeper understanding of the context. To achieve this, we employ the MCTS-inspired search method to explore the knowledge base and locate the relevant subregion.

In each selection step, the candidates for knowledge sentences are chosen from the semantic group associated with the current concept node. The policy model assesses the probability of selecting each candidate based on their cosine similarity to both the conversation context and explicit concepts mentioned in the context.

The critic model, denoted as $c_{bridge}$, evaluates the selected knowledge sentences based on two metrics: the concept bridging rate and the semantic coherence score. The concept bridging rate measures the proportion of explicit context concepts ${N}_{C}$ that are found within the predicted concept node's semantic group $G({N}_{P})$. The semantic coherence score is calculated using the Wasserstein Distance ($\textrm{WD}$) ~\cite{panaretos2019statistical} between the conversation context $C$ and the selected knowledge sentence $k$.

\begin{equation}
\setlength{\abovedisplayskip}{5pt}
\setlength{\belowdisplayskip}{5pt}
    c_{brdige} = |{N}_{C} \cap G({N}_{P})| / |{N}_{C}| + \lambda \textrm{WD}(x, C)
\end{equation}
where $|\cdot|$ denotes the number of elements in a set. $\lambda$ is a constant. By using this composite score, the critic model can provide a comprehensive evaluation of the selected knowledge sentences, guiding the policy model to make informed decisions in the concept bridging process.

\subsection{Reasoning-Aware Knowledge Retrieval Module}

The Reasoning-Aware Knowledge Retrieval module is designed to retrieve knowledge sentences that provide evidence, facts, or explanations to support logical inferences. Building upon the Concept Bridging step, this module leverages the MCTS-inspired search method to thoroughly explore the search space and identify the most relevant knowledge sentences for each logical inference. By doing so, the module aims to provide a robust and informed foundation for logical reasoning and decision-making.

In the Reasoning-Aware Knowledge Retrieval module, we combine the logical inference and conversation context into a single input to facilitate a comprehensive understanding of the context. The policy model then evaluates the probability of each candidate knowledge sentence by assessing its semantic similarity with this input, ensuring that the selected knowledge is relevant to both the logical inference and conversation context.
The critic model, $c_{retrieve}$, plays a crucial role in evaluating the selected knowledge sentences. It utilizes the DSE model \cite{Zhou2022} to assess the appropriateness of the selected knowledge $k$ for both the logical inference $r$ and the context $C$. Additionally, the critic model rewards knowledge sentences that provide more information, assuming that longer knowledge sentences contain more valuable insights. Conversely, it penalizes semantic repetition among the retrieved knowledge to avoid redundant information.

The critic model $c_{retrieve}$ is defined as:
\begin{equation}
    c_{retrieve} = sim(r, k) + sim(C, k)  + |k| - sim(k, \hat k)
\end{equation}
where $sim$ represents the cosine similarity measurement, $|\cdot|$ denotes the knowledge length, $k$ is the knowledge retrieved in the current step, and $\hat k$ is the knowledge retrieved from previous steps. By optimizing this composite score, the critic model guides the policy model to select knowledge sentences that are informative, relevant, and non-redundant, ultimately enhancing the reasoning-aware knowledge retrieval process.

\section{Experiments}
In this section, we conduct experiments on two representing multi-turn dialogue datasets to show the effectiveness of our method. We mainly focus on the following questions:

\begin{itemize}[leftmargin=*]
    \item To what extent is our retrieved reasoning-aware knowledge semantically consistent with the knowledge inferred by LLM?

    \item Does our retrieved reasoning-aware knowledge exhibit higher pairwise diversity?
    
    \item Does our retrieved reasoning-aware knowledge align better with the logical structure found in human conversations?

    \item Does our retrieved reasoning-aware knowledge contribute to better conversation responses?


\end{itemize}

\subsection{Implementation Details}

We implemented our algorithm using PyTorch and tested it on an NVIDIA V100 GPU with 16 GB of memory. Below are the details for the MCTS and Critic Model:

In the MCTS-inspired knowledge base exploration process, the maximum path length we traverse—referred to as the search horizon—is limited to 3. This means we explore up to three steps from the starting point to uncover relevant knowledge. The exploration involves using 50 knowledge sentences that have been pruned with a similarity function to reduce the search space. Additionally, each action produces a return of 5 knowledge sentences.
In the PUCT algorithm, the exploration constant, denoted as $c_{\textrm{puct}}$ is set to $1 / \sqrt2$. This value balances exploration and exploitation during the traversal of the knowledge graph.

To measure the semantic coherence between the context and the candidate knowledge, we use the Wasserstein Distance constant, represented as $\lambda$. By setting $\lambda$ to -1000, we scale the distance to a range between 0 and 1. We also introduce a factor that encourages the inclusion of longer knowledge facts in the search results. To account for this, we add the weighted length of the knowledge fact to the value function, with the weight assigned to length set at 0.1.

\subsection{Experiments Setup}
\subsubsection{\textbf{Datasets}} 

We evaluate our approach using the DailyDialog (DD) \cite{li2017dailydialog} and Empathetic Dialogue (EMP) \cite{rashkin2019towards} datasets, which comprise multi-turn open-domain dialogues reflecting daily communication scenarios. We use the entire test set including 1000 dialogues for DD and 2545 dialogues for EMP.
For knowledge retrieval, we utilize the GenericsKB Best corpus, which consists of 1,025,413 distinct commonsense knowledge facts.

\subsubsection{\textbf{Baselines}} We assess the effectiveness of our reasoning-aware knowledge retrieval method by benchmarking it against the typical RAG framework. RAG framework involves a two-stage process, where candidate sentences are first retrieved from a vector database based on embedding similarity search, and then ranked using a reranker model. We employ five distinct embedding methods to facilitate a comprehensive comparison.
SBERT \cite{reimers2019sentence} encodes sentences into BERT embeddings for semantic matching.
DPR \cite{karpukhin2020dense} utilizes a dense vector index and a bi-encoder architecture for efficient retrieval.
Contriever \cite{izacard2022unsupervised} uses a contrastive learning framework to pre-train models for knowledge retrieval. 
GTE \cite{li2023towards} improves Contriever by training general-purpose text embeddings using multi-stage contrastive learning.
Instructor \cite{su-etal-2023-one}, an instruction-finetuned text embedding model that adjusts embeddings to suit different knowledge retrieval tasks.

To ensure a fair comparison, we input the concatenation of the conversation context and the logical inferences obtained from our Reasoner module into these baseline models. 

\subsubsection{\textbf{Ablation Methods}} 

To gain a deeper understanding of the contributions of each module to our method's effectiveness, we conduct two ablation studies. These studies help us identify the most impactful module and quantify its contribution to the overall performance.
In the first ablation study \textbf{w/o Reasoner}, we exclude the Reasoner module from our pipeline. This means that we only use the conversation context as input to perform our MCTS-inspired knowledge retrieval. By doing so, we can assess the importance of the Reasoner module in guiding the knowledge retrieval process.
In the second ablation study \textbf{w/o Concept Bridging}, we remove the Concept Bridging module, which focuses on retrieving context-relevant knowledge sentences. Instead, we retrieve the entire knowledge base, without any filtering or prioritization. This study allows us to evaluate the effectiveness of the Concept Bridging module in identifying the most relevant knowledge sentences for a given conversation context.

\begin{table*}[hbtp]
\caption{Evaluation Results of Semantic Similarity compared with LLM Inferences. }
     
\footnotesize
\centering

\renewcommand\arraystretch{1.1}
\resizebox{0.95\textwidth}{!}{%
\begin{tabular}{p{0.5cm}lcccccccccc}
\toprule
\multirow{2}{*}{Data} & \multirow{2}{*}{Method} & \multicolumn{6}{c}{BERTScore $\uparrow$ } & \multicolumn{2}{c}{BARTScore $\uparrow$} & \multicolumn{2}{c}{MoverScore $\uparrow$} \\

& & avg P & avg R & avg F & max P & max R & max F & avg & max & avg & max \\                                                            
\cmidrule(r){1-2} \cmidrule(r){3-8} \cmidrule(r){9-10} \cmidrule(r){11-12}

\multirow{8}{*}{\begin{tabular}[c]{@{}l@{}}DD\end{tabular}}      
&SBERT & 80.25  & 80.77  & 80.50  & 82.10 & 82.64 & 82.16 & -4.94 & -3.73 & 76.91 & 79.62 \\ 
& DPR            &   83.62    &    84.72   &   84.15    &    84.51   &   86.25    &   85.22    &   -4.39  &  -3.64   &  80.06   &   81.74    \\
& Contriever            &   83.38    &    84.07   &   83.71    &   85.74    &  86.81     &   86.11    &   -4.15  &  -3.28   &  80.24   &   81.98 \\
& GTE            &   84.07    &    84.50   &   84.13    &   86.07    &  86.50     &   86.13    &   -4.09  &  -3.20   &  80.26   &   82.53 \\
& Instructor            &   84.22    &    84.10   &   84.19    &   86.23    &  86.48     &   86.19    &   -4.03  &  -3.21   &  80.29   &   82.54 \\
\cline{2-12}
& Ours & \textbf{84.29} & \textbf{85.05} & \textbf{84.65} & \textbf{89.10} & \textbf{89.43} & \textbf{88.90} & -4.19 & \textbf{-2.85} & \textbf{80.32} & \textbf{83.78} \\
&  w/o Reasoner & 84.23 & 85.04 & 84.62 & 88.19 & 88.91 & 88.31 & \textbf{-4.02} & -2.98 & 80.15 & 83.06 \\ 
& w/o Concept Bridging &   80.76    &    84.79   &   82.73    &   81.92    &  85.63     &  83.57     &  -4.63   &  -4.27   &  77.23   &  78.08   \\

\hline
\multirow{8}{*}{\begin{tabular}[c]{@{}l@{}}EMP\end{tabular}} 
& SBERT  & 80.72 & 80.83 & 80.77  &  82.65  & 83.12 & 82.88 & -4.56 & -3.38 & 75.36 & 78.49 \\
& DPR            &   83.95    &    85.29   &   84.60    &   84.83    &  86.73     &   85.62    &   -4.07  &  -3.38   &  79.61   &   81.09 \\
& Contriever            &   83.50    &    84.43   &   83.96    &   87.32    &  88.12     &   87.97    &   -4.05  &  -3.22   &  78.43   &   81.87 \\
& GTE            &   84.12    &    84.65   &   84.30    &   88.17    &  88.67     &   88.31    &   -3.93  &  -2.90   &  78.89   &   82.41 \\
& Instructor            &   84.84    &    85.26   &   85.11    &   88.49    &  89.04     &   88.73    &   -3.77  &  -2.82   &  79.60   &   82.48 \\
\cline{2-12}
& Ours & \textbf{85.27} & \textbf{86.54} & \textbf{85.89} & \textbf{88.93} & \textbf{89.59} & \textbf{88.90} & \textbf{-3.71} & \textbf{-2.78} & \textbf{79.99} & \textbf{83.00} \\
& w/o Reasoner  & 84.78 & 85.77 & 85.26 & 87.80 & 89.07 & 88.25 & -3.92 & -2.93 & 79.82 &  82.05 \\ 
& w/o Concept Bridging &   80.21    &   85.35    &   82.74    &  81.43    &   85.98   &   83.32   &  -4.39  &  -3.87   &  77.32   &  79.01  \\
\bottomrule

\end{tabular}%
}
\label{tab:similarity chatgpt}
\end{table*}

\subsection{Similarity with LLM Inferences}

As LLMs have demonstrated exceptional performance on commonsense knowledge tasks, we compare our retrieved reasoning-aware knowledge with the inferences generated by LLMs. 
We utilize ChatGPT to elicit relevant knowledge by providing the following prompt: ``Given a conversation between two people, what commonsense knowledge could you infer from the conversation?" By doing so, we can tap into ChatGPT's internal knowledge base and compare its performance with our reasoning-aware knowledge retrieval method.
We evaluate the effectiveness of our retrieved reasoning-aware knowledge by comparing its similarity to ChatGPT inferences, thereby assessing how well it captures relevant knowledge relative to a strong LLM.

\subsubsection{\textbf{Evaluation Metrics} }
We employ BERTScore \cite{zhangbertscore}, BARTScore \cite{yuan2021bartscore}, and MoverScore \cite{zhao2019moverscore} as semantic similarity measurements. 
We calculate both the average and maximum values for each evaluation metric. The average value assesses the overall semantic similarity and the maximum value focuses on the highest similarity between the retrieved knowledge and one of ChatGPT-inferred knowledge sentences.

\subsubsection{\textbf{Results and Analysis}} 

Our retrieved reasoning-aware knowledge demonstrates superior performance compared to previous RAG baselines on both datasets, as shown in Table \ref{tab:similarity chatgpt}. The impressive scores achieved in BERTScore, BARTScore, and MoverScore indicate a strong semantic consistency with the knowledge inferred by ChatGPT, suggesting an excellent overall quality of the retrieved knowledge. Notably, the increase in the maximum score suggests that our knowledge retrieval method is effective in identifying a knowledge sentence that closely resembles the one inferred by ChatGPT, further highlighting its capabilities.

\subsubsection{\textbf{Ablation Study}} 
Through our ablation studies, we observe that removing concept bridging leads to a significant decrease in LLM inference similarity, as retrieving the entire knowledge base leads to inaccurate knowledge retrieval. 
After removing the Reasoner module, we observe a slight decrease in the similarity between our retrieved reasoning-aware knowledge and ChatGPT inferences, suggesting that while ChatGPT's inferences do exhibit some context-specific logical reasoning, they fall short of the more comprehensive logical reasoning capabilities that the Reasoner module provides.

\begin{table*}[]
\caption{Results of Knowledge Pair-Wise Diversity. Higher diversity indicates lower Semantic and Token Overlap. }
\centering
\small
\renewcommand\arraystretch{0.95}
\newcolumntype{C}[1]{>{\centering\arraybackslash}p{#1}}
\resizebox{\textwidth}{!}{%
\begin{tabular}{llC{1.7cm}C{1.7cm}C{1.7cm}C{1.7cm}C{1.7cm}C{1.7cm}}
\toprule
\multirow{2}{*}{Data} & \multirow{2}{*}{Method} & \multicolumn{3}{c}{Semantic Overlap $\downarrow$} & \multicolumn{3}{c}{Token Overlap $\downarrow$} \\
\cmidrule(r){3-5} \cmidrule(r){6-8}
                         &                         & Precision          & Recall          & F1 Score          & Rouge 1        & Rouge 2        & Rouge L        \\
\cmidrule(r){1-2} \cmidrule(r){3-5} \cmidrule(r){6-8}
\multirow{9}{*}{DD}      
& SBERT                   &     88.85       &      88.84      &      88.83      &     30.43      &      12.99     &      28.06     \\
& DPR   &   86.75        &      86.79      &    86.75        &      18.05      &     4.70      &      17.31                \\
& Contriever   &   88.62        &      88.45      &    88.51        &      25.28      &     9.33      &      22.95                \\
& GTE   &   89.57        &      89.54      &    89.53        &      28.49      &     11.33      &      26.73                \\
& Instructor   &   89.58        &      89.76      &    89.62        &      27.97      &     13.88      &      26.76                \\
& ChatGPT Inference       &     88.38       &      87.85      &       88.11    &      22.82     &     6.03      &     19.35      \\\cline{2-8}
& Ours           &     \textbf{83.63}       &     \textbf{83.68}       &       \textbf{83.64}     &     \textbf{10.29}     &     \textbf{1.14}      &     \textbf{9.16}      \\
& w/o Reasoner   &      83.77      &     83.71       &       83.72     &     13.22      &     2.22      &      11.72     \\
& w/o Concept Bridging    &      86.37      &       86.39     &       86.38     &     21.94      &     6.65      &     17.84      \\
\hline
\multirow{9}{*}{EMP}     & SBERT                   &      89.32      &      89.32      &       89.32     &     28.73      &     10.88      &      26.41     \\
& DPR              &      86.81      &    86.67        &     86.72       &      14.39     &     2.94      &    13.53       \\
& Contriever   &   88.93        &      88.62      &    88.77        &      23.64      &     7.36      &      21.25                \\
& GTE   &   89.76        &      89.49      &    89.58        &      26.17      &     8.49      &      23.60                \\
& Instructor   &   89.28        &      89.46      &    89.39        &      23.80      &     8.16      &      22.36                \\
& ChatGPT Inference       &      88.97      &     88.55       &       88.75     &     21.04      &     4.61      &     18.10      \\\cline{2-8}
& Ours           &      \textbf{84.71}      &      \textbf{84.72}      &       \textbf{84.70}     &      \textbf{9.64}     &     \textbf{1.05}      &      \textbf{8.66}     \\
& w/o Reasoner   &       85.09     &      85.09      &      85.08      &      10.35     &     1.27      &  9.21  \\
& w/o Concept Bridging    &     86.53       &   86.26         &      86.41      &    13.29       &    2.76       &     11.83      \\\bottomrule
\end{tabular}
}
\label{tab:diverse and event cover}
\end{table*}

\subsection{Pair-Wise Diversity}

We examine the diversity of knowledge sentences generated by each approach, comparing the performance of traditional RAG methods, our reasoning-aware knowledge retrieval method, and the prompt-based ChatGPT Inference method. To quantify their diversity, we calculate the average pairwise difference between the sentences produced by each method.

\subsubsection{\textbf{Evaluation Metrics}} 

To evaluate the diversity among knowledge sentences, we employ two metrics:
BERTScore \cite{zhangbertscore} measures the semantic overlap between each pair of knowledge sentences, capturing their similarity in meaning.
ROUGE score \cite{lin2004rouge} quantifies the token overlap between the sentences, counting the number of shared tokens.
A higher diversity score indicates that the method has produced knowledge sentences that cover a broader range of information and viewpoints.

\subsubsection{\textbf{Results and Analysis}} 
The results of our retrieved knowledge pair-wise diversity evaluation are presented in Table \ref{tab:diverse and event cover}. Our analysis reveals that our reasoning-aware knowledge retrieval method surpasses both previous RAG methods and the ChatGPT Inference method in terms of pair-wise knowledge diversity. This is evident from the lower token overlap and lower semantic overlap observed among the knowledge sentences generated by our method.
It suggests that, while ChatGPT is proficient in generating relevant knowledge, it has limitations when it comes to exploring diverse aspects of a given conversation. 
In contrast, by incorporating logical inferences, our reasoning-aware retrieval method is able to identify knowledge that encompasses a broader range of aspects.

\subsubsection{\textbf{Ablation Study}} 
Notably, our knowledge retrieval method is still able to uncover a more diverse range of knowledge compared to previous RAG methods, even when the Reasoner and Concept Bridging modules are removed. This underscores the significance of our iterative MCTS-inspired search approach. The critic model's reward function, which favors knowledge that provides more information and discourages semantic repetition, enables the MCTS method to retrieve knowledge with minimal sentence-level repetition, even in the absence of the Reasoner module.

\subsection{Human Logic Alignment}

We further examine how well our reasoning-aware knowledge aligns with the logical structure that underlies human conversations. To uncover this structure, we identify and annotate the key events that drive the conversation's overall logical flow. We then ask ChatGPT to label each element within this structure, referred to as a logic transition, using the following instruction: ``Given a conversation between two people, please label the phrases that could guide the development of the conversation."

\subsubsection{\textbf{Evaluation Metrics}} 
We evaluate the compatibility between the retrieved knowledge and the logical structure found in human conversations, as outlined in the ACCENT framework \cite{ghazarian-etal-2023-accent}. 
Specifically, we adopt the ACCENT approach, which considers the events inferred by the event-centric knowledge graph as ground truth labels, denoted as $E$. We consider a successful human logic mapping to occur when both the retrieved knowledge and the human logic transition align with the ground-truth event inferred by the knowledge graph. To quantify this alignment, we set a threshold $\theta$ to measure entailment. The human logic alignment score, $S_{align}$, is then calculated as the ratio of the intersection of the retrieved knowledge and human logic transitions that exceed the threshold $\theta$ to the total number of human logic transitions that exceed the threshold $\theta$. This score provides a quantitative measure of how well the retrieved knowledge maps to the human logic transitions.
\begin{equation}
\label{Eq:align}
S_{align} = \frac{|{K}_{\theta} \cap T_{\theta}|}{|T_{\theta}|}, \left\{
\begin{aligned}
 & P_{e}(k|E)>\theta, k\in K_{\theta}\\
 & P_{e}(T|E)>\theta, t\in T_{\theta}.
\end{aligned}
\right.
\end{equation}
The number of elements in a set is denoted by $|\cdot|$. Meanwhile, $P_{e}$ represents the entailment model \footnote{https://huggingface.co/facebook/bart-large-mnli}. We label the retrieved knowledge and logic transitions with entailment scores that surpass the threshold $\theta$ as ${K}_{\theta}$ and ${T}_{\theta}$, respectively.

\begin{table}[htbp]
\caption{Results of Alignment with Human Logic.}
\resizebox{0.46\textwidth}{!}{%
\begin{tabular}{lcc}
\toprule
\multicolumn{3}{>{\centering\cellcolor{gray!20}}c}{DailyDialog} \\
\multirow{2}{*}{Method} & \multicolumn{2}{c}{Human Logic Alignment} \\
                        & Threshold = 0.5       & no Threshold      \\
\hline
SBERT                   & 38.46                 & 63.53             \\
DPR                     & 42.12                 & 67.35             \\
Contriever                     & 41.10                 & 66.68             \\
GTE                     & 43.41                 & 62.37             \\
Instructor                     & 45.56                 & 66.98             \\
ChatGPT Inference       & 55.82                 & 75.47             \\
\hline
Ours                    & \textbf{80.97}                 & \textbf{95.15}             \\
w/o Reasoner            & 57.06                 & 73.76             \\
w/o Concept Bridging    & 42.17                   &  68.39              \\
\hline
\multicolumn{3}{>{\centering\cellcolor{gray!20}}c}{Empathetic Dialogue} \\
\multirow{2}{*}{Method} & \multicolumn{2}{c}{Human Logic Alignment} \\
                        & Threshold = 0.5       & no Threshold      \\
                        \hline
SBERT                   & 39.55                 & 57.43             \\
DPR                     & 33.99                 & 62.83             \\
Contriever                     & 37.41                 & 59.28             \\
GTE                     & 42.32                 & 61.85             \\
Instructor                     & 44.97                 & 65.62             \\
ChatGPT Inference       & 55.43                 & 68.17             \\
\hline
Ours                    & \textbf{85.20}                 & \textbf{96.02}             \\
w/o Reasoner            & 53.93                 & 69.49             \\
w/o Concept Bridging    &  39.82                     &       63.45           \\
\bottomrule
\end{tabular}
}
\label{tab:human logic cover}
\end{table}

\subsubsection{\textbf{Results and Analysis}} 
As shown in Table \ref{tab:human logic cover}, our analysis reveals that previous RAG methods struggle to retrieve knowledge that aligns with human logic, succeeding in only 67\% of cases. This limitation suggests that these methods are unable to effectively support both logical inferences and conversation context.
In contrast, ChatGPT Inferences demonstrate a stronger performance, covering 75\% of human events in the conversation. While ChatGPT excels in terms of relevance, it falls short in actively reasoning about the underlying commonsense knowledge that aligns with human thinking.
Notably, our reasoning-aware knowledge retrieval method shows a significant boost in human logic alignment score. This improvement confirms our method's ability to successfully capture the underlying reasoning-aware knowledge that aligns with human logic.

\subsubsection{\textbf{Ablation Study}}
Our ablation study reveals the significance of each module in our reasoning-aware knowledge retrieval method. Notably, when we remove the Reasoner module, the human logic alignment score drops to the level of ChatGPT's inferred knowledge, indicating that the Reasoner module plays a crucial role in capturing logical inference-coherence.
Meanwhile, excluding the Concept Bridging module causes the alignment score to drop to the level of previous RAG methods, emphasizing the importance of Concept Bridging in identifying context-coherent knowledge.
These results underscore the challenges of retrieving knowledge that satisfies both context-coherence and logical inference-coherence solely through embedding similarity concatenation. Our approach, which incorporates Concept Bridging to focus on relevant knowledge and the Reasoner module to ensure logical inference-coherence, addresses these challenges and achieves improved performance.

\begin{figure}[htbp]
\centering
\includegraphics[width=0.48\textwidth]{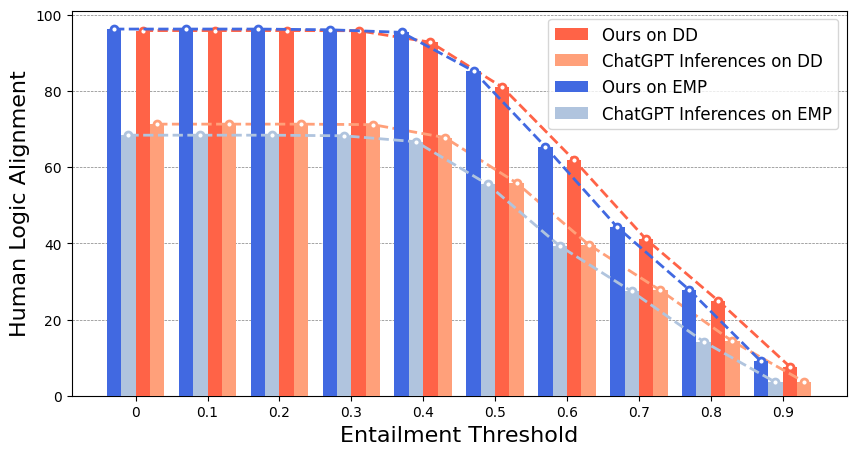}
\caption{Comparison of human logic alignment coverage between our method and ChatGPT-inferred knowledge across varying entailment thresholds.}
\label{figure:coverage-threshold}
\end{figure}

\subsubsection{\textbf{Threshold Analysis}} 
We further explore the effect of entailment thresholds on the percentage of human logic alignment in Figure \ref{figure:coverage-threshold}. Our analysis shows that our method consistently surpasses ChatGPT-inferred knowledge in capturing human logic transitions across the entire range of entailment thresholds, from 0 to 1.
Notably, as the entailment threshold is increased, the performance gap between our method and ChatGPT-inferred knowledge gradually narrows, shrinking from a 20\% advantage to near parity. This trend not only validates the robustness of our evaluation methodology but also underscores the effectiveness of our retrieval approach in capturing human logic transitions.

\subsection{Response Quality Evaluation}

In our response evaluation experiment, we investigate whether our reasoning-aware knowledge retrieval method can improve the quality of responses generated by ChatGPT. To do this, we conduct a series of evaluations where ChatGPT generates responses based on conversation prompts that incorporate retrieved knowledge as a reference point. Specifically, we prompt ChatGPT with the instruction: ``Please generate the response for this conversation, you can consider the knowledge of XXX for your reference in your response generation."

We compare the responses generated by ChatGPT that reference our retrieved knowledge against two other types of responses:
1) Original ChatGPT Responses: These responses are generated by ChatGPT solely based on the given conversation context, without any additional knowledge integration.
2) ChatGPT Inference Responses: In this case, ChatGPT incorporates its own commonsense knowledge inference to produce responses, allowing it to leverage its internal understanding of the conversation's context.

\begin{table}[htbp]
\caption{Results of Response Quality Evaluation. }
\renewcommand\arraystretch{1.0}
\resizebox{0.48\textwidth}{!}{%
\begin{tabular}{
  l
  >{\centering\arraybackslash}p{1.1cm}
  >{\centering\arraybackslash}p{1.1cm}
  >{\centering\arraybackslash}p{1.1cm}
  >{\centering\arraybackslash}p{1.1cm}
}
\toprule
\multicolumn{5}{>{\centering\cellcolor{gray!20}}c}{DailyDialog} \\
& Coherent & Informative & Creative & Logic \\
\hline
ChatGPT  &     \textbf{4.09}     &     3.65        &     2.56     &    2.62   \\
w/ Inference &    3.89      &      3.66       &    3.12      &   2.85    \\
Ours     &     4.08     &      \textbf{4.01}       &      \textbf{3.59}    &  \textbf{3.19}  \\
\hline
\multicolumn{5}{>{\centering\cellcolor{gray!20}}c}{Empathetic Dialogue} \\
& Coherent & Informative & Creative & Logic \\
\hline
ChatGPT  &     \textbf{4.02}     &      3.51       &     2.52     &   2.61    \\
w/ Inference &    3.84      &       3.58      &    3.14      &    2.75   \\
Ours     &     3.98     &      \textbf{3.82}       &     \textbf{3.27}     &  \textbf{3.01}  \\
\bottomrule
\end{tabular}
}
\label{tab:response eval}
\end{table}

\subsubsection{\textbf{Evaluation Metrics}} We adopt a hybrid evaluation approach that combines LLM-based and human judgments to assess the quality of dialogue responses \cite{gao2025llm, liu2023g, wang2023chatgpt}. Given the complexity and subjectivity involved in scoring dialogue responses across multiple dimensions (e.g., coherence, relevance, diversity), we first use ChatGPT to generate preliminary scores — leveraging its demonstrated ability to produce evaluations that align closely with human judgments \cite{wang2023chatgpt}. Recent studies have shown that LLM-based evaluators like ChatGPT can achieve high correlation with human annotators, particularly on structured tasks such as dialogue fluency and logical consistency \cite{liu2023g}, and are often more consistent and scalable than individual human raters \cite{gao2025llm}.

To ensure reliability and validity, these automated scores are then reviewed and refined by three expert human annotators from a mutual assistance platform. This hybrid design allows us to benefit from both the efficiency of LLM-based scoring and the nuanced judgment of human experts, ensuring high-quality and well-grounded evaluation outcomes.
During the review process, human evaluators are instructed to carefully assess the responses based on criteria such as coherence, informativeness, creativity, and logical consistency. They are responsible for making any necessary adjustments to the scores assigned by ChatGPT, ensuring a thorough evaluation of the dialogue responses.

\begin{table}[htbp]
\caption{Results of Multiple Response Quality Evaluation. }
\small
\renewcommand\arraystretch{1.0}
\resizebox{0.47\textwidth}{!}{%
\begin{tabular}{
  l
  >{\centering\arraybackslash}p{3cm}
  >{\centering\arraybackslash}p{3cm}
}
\toprule
\multicolumn{3}{>{\centering\cellcolor{gray!20}}c}{DailyDialog} \\
& Pairwise Diversity & Knowledge Depth \\
\hline
ChatGPT  &     9.5\%     &     11.0\%           \\
w/ Inference &    12.0\%      &      23.5\%          \\
Ours     &     \textbf{78.5\%}     &      \textbf{65.5\%}        \\
\hline
\multicolumn{3}{>{\centering\cellcolor{gray!20}}c}{Empathetic Dialogue} \\
& Pairwise Diversity & Knowledge Depth  \\
\hline
ChatGPT  &     10.5\%     &      12.0\%        \\
w/ Inference &    14.5\%      &       32.0\%      \\
Ours     &     \textbf{75.0\%}     &      \textbf{56.0\%}       \\
\bottomrule
\end{tabular}
}
\label{tab:response_eval_2}
\end{table}

Moreover, to evaluate the quality of multiple dialogue responses generated by different methods, evaluators use a win-loss approach. This involves comparing the responses in pairs and determining which method produces the better response based on quality. The evaluation focuses on two key aspects: pairwise diversity, which assesses the variety of responses generated, and knowledge depth, which evaluates the richness and comprehensiveness of the information provided in the responses.

\subsubsection{\textbf{Results and Analysis}}

The results shown in Table \ref{tab:response eval} indicate that when ChatGPT leverages our retrieved knowledge, it achieves the highest scores in informativeness, creativity, and logical complexity. However, coherence scores show a slight decrease, likely because longer responses tend to receive lower ratings compared to shorter, more concise ones. 
The qualitative and quantitative analysis of multiple reasoning paths is summarized in Table \ref{tab:response_eval_2}. 
Within our MCTS-based generation framework, each response is derived from a distinct knowledge path explored during the search process. This design enables us to assess how variations in knowledge selection influence the final outputs in terms of response diversity and knowledge depth. We find strong alignment between human judgments of response diversity and our automated metric of knowledge pairwise diversity, validating the effectiveness of our diversity measurement. Moreover, our retrieval method consistently produces more comprehensive and knowledge-rich responses.


\section{Conclusion}
In this study, we introduce a novel reasoning-aware commonsense knowledge retrieval method designed specifically for conversational agents. Our approach frames the knowledge retrieval process as a multi-objective optimization (MOO) problem, balancing coherence with the conversation context and alignment with logical reasoning outcomes. Additionally, we employ a Monte Carlo Tree Search (MCTS)-inspired method that effectively navigates the knowledge base, uncovering complex interconnections between sentences. Experiments conducted on two representative multi-turn dialogue datasets reveal that our retrieved knowledge aligns more closely with the logical structure of human conversations, resulting in responses that are both more informative and creative. Looking ahead, future research could focus on developing advanced logical reasoning modules to enhance performance in open-domain text generation tasks.

\bibliographystyle{IEEEtran}
\bibliography{custom}

\end{document}